\documentclass{bmvc2k}

\usepackage{amssymb}
\usepackage[ruled,vlined]{algorithm2e}
\usepackage{multirow}
\usepackage{booktabs}
\usepackage{diagbox}
\usepackage{floatrow}
\newfloatcommand{capbtabbox}{table}[][\Xhsize]

\newcommand{\Lagr}{\mathcal{L}}

\DeclareMathOperator*{\argmin}{argmin}

\title{Generative Dynamic Patch Attack}

\addauthor{Xiang Li}{xli62@gsu.edu}{1}
\addauthor{Shihao Ji}{sji@gsu.edu}{1}

\addinstitution{
 Department of Computer Science\\
 Georgia State University\\
 Atlanta, GA, USA
}

\runninghead{Xiang Li, Shihao Ji}{Generative Dynamic Patch Attack}

\begin{document}

\maketitle
\vspace{-10pt}
\begin{abstract}
Adversarial patch attack is a family of attack algorithms that perturb a part of image to fool a deep neural network model. Existing patch attacks mostly consider injecting adversarial patches at input-agnostic locations: either a predefined location or a random location. This attack setup may be sufficient for attack but has considerable limitations when using it for adversarial training. Thus, robust models trained with existing patch attacks cannot effectively defend other adversarial attacks. In this paper, we first propose an end-to-end patch attack algorithm, Generative Dynamic Patch Attack (GDPA), which generates both patch pattern and patch location adversarially for each input image. We show that GDPA is a generic attack framework that can produce dynamic/static and visible/invisible patches with a few configuration changes. Secondly, GDPA can be readily integrated for adversarial training to improve model robustness to various adversarial attacks. Extensive experiments on VGGFace, Traffic Sign and ImageNet show that GDPA achieves higher attack success rates than state-of-the-art patch attacks, while adversarially trained model with GDPA demonstrates superior robustness to adversarial patch attacks than competing methods. 
Our source code can be found at \url{https://github.com/lxuniverse/gdpa}.
\end{abstract}

\vspace{-10pt}
\section{Introduction}
\vspace{-15pt}
\begin{figure}[ht!]
\centering
\includegraphics[width=0.8\textwidth]{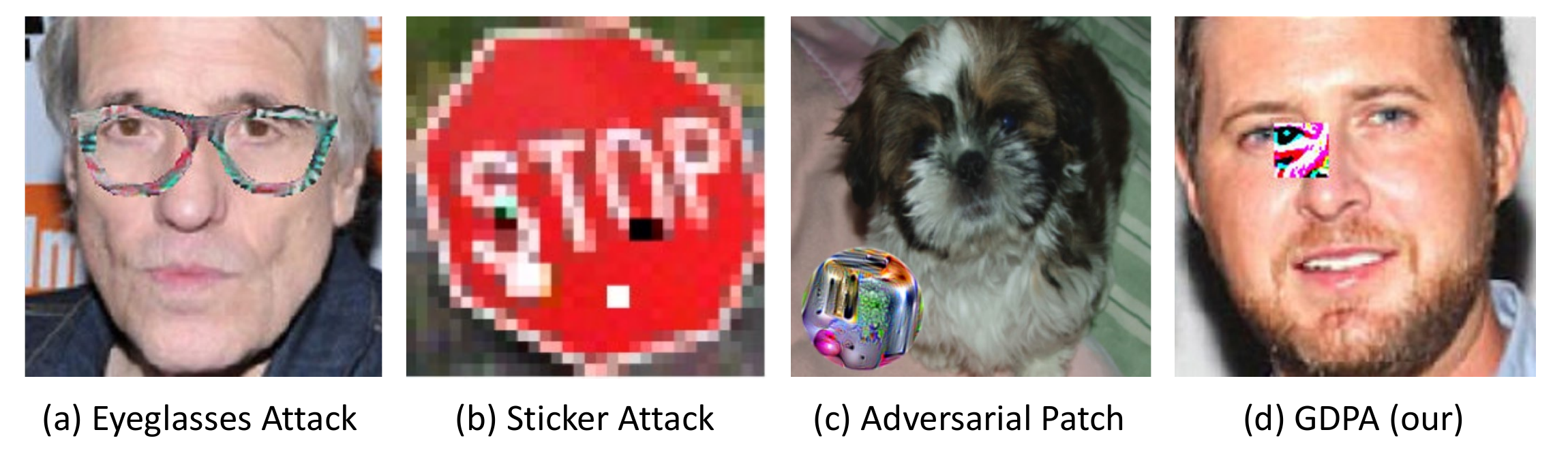}
\vspace{-5pt}
\caption{Different types of patch attacks: 
(a) Eyeglasses Attack~\cite{sharif2016accessorize}, (b) Sticker Attack~\cite{evtimov2017robust}, (c) Adversarial Patch~\cite{brown2017adversarial}, and (d) GDPA (ours).}
\label{fig:examples}
\vspace{-10pt}
\end{figure}

Deep neural networks (DNNs) have demonstrated remarkable success in solving complex prediction tasks in a variety of fields: computer vision~\cite{deng2009imagenet}, natural language processing~\cite{sutskever2014sequence} and speech recognition~\cite{senior2012deep}. However, recent studies show that they are particularly vulnerable to adversarial examples~\cite{goodfellow2014explaining} in the form of small perturbations to inputs that lead DNNs to predict incorrect outputs. 

Recent works~\cite{brown2017adversarial, karmon2018lavan, evtimov2017robust, sharif2016accessorize, yang2020patchattack, yang2019design}, show that perturbing part of an image with perceivable noise is another effective method to attack neural network models. Typically, attackers can craft perceivable patches to replace part of images for adversarial attack. The advantage of this perceivable patch attack is that it is more practical than the imperceptible adversarial attacks in the real world: adversaries can paste a sticker on a traffic sign to attack the autopilot system of autonomous vehicles. There are several situations where patch attack is significant concerning due to its security threats: 1) an attacker uses adversarially designed eyeglass frames~\cite{sharif2016accessorize} to fool face recognition (Fig.~\ref{fig:examples}a), 2) an attacker pastes adversarially crafted stickers~\cite{evtimov2017robust} on stop signs to fool traffic sign classification (Fig.~\ref{fig:examples}b), and 3) a universal adversarial patch~\cite{brown2017adversarial} causes targeted misclassification of any object (Fig.~\ref{fig:examples}c).

However, it is a significant limitation that most patch attack algorithms do not consider the problem of finding the best location in an image to inject the patch. Existing patch attack algorithms either use a fixed position as patch location~\cite{sharif2016accessorize,evtimov2017robust,yang2019design} or learn patches that are universal across different locations~\cite{brown2017adversarial,karmon2018lavan,yang2020patchattack}. The fixed location methods show high attack success rates but are poorly performed at other locations, while the random location patches do not have competitive attack success rates compared to the fixed location methods. To address this issue, in this paper we propose a Generative Dynamic Patch Attack (GDPA), which learns image-dependent patch pattern and patch location altogether. GDPA is inspired by the idea that different images have different sets of weak pixels since DNN classifiers typically focus on different image regions when queried by different images~\cite{simonyan2013deep}. Therefore, an image-dependent dynamic patch attack would be more effective than a fixed location or random location patch attack.

On the other hand, due to the security threats of adversarial attacks, a variety of adversarial defense algorithms have been developed recently~\cite{goodfellow2014explaining,lyu2015unified,shaham2018understanding}, among which adversarial training (AT)~\cite{goodfellow2014explaining} has been proved the most effective one for hardening neural networks against adversarial attacks. Although AT with the PGD attack~\cite{madry2017towards} is the most scalable and effective method for learning robust models, a recent work of Wu et al.~\cite{wu2019defending} shows that AT exhibits limited effectiveness against three high-profile physically realizable patch attacks: eyeglasses attack~\cite{sharif2016accessorize}, sticker attack~\cite{evtimov2017robust} and adversarial patch~\cite{brown2017adversarial}. To overcome this limitation, Wu et al.~\cite{wu2019defending} propose a Rectangular Occlusion Attack (ROA) for adversarial training, which yields models highly robust to patch attacks. ROA is a two-stage patch attack algorithm, which first uses a \emph{gray pattern} to find the location in image that maximizes the cross-entropy loss via grid search, and then optimizes the patch pattern at the identified position. However, this two-stage patch attack method is suboptimal and has quite a few limitations (see a discussion in Sec.~\ref{sec:related}), which motivates us to propose GDPA that learns patch pattern and patch location simultaneously. Moreover, to improve the inference efficiency, GDPA employs a generator to generate patch pattern and location with one forward propagation, without expensive iterative optimizations that are employed by other attack algorithms, such as PGD~\cite{madry2017towards} and ROA~\cite{wu2019defending}. Concretely, we make the following contributions: 
\begin{itemize}
    \item We introduce a generic patch attack method GDPA that can generate dynamic/static and visible/invisible patch attacks with a few configuration changes. 
    \item GDPA employs a generator to generate patch pattern and patch location altogether per image, and reduces the inference time substantially (e.g., 40-50x faster).
    \item GDPA is an end-to-end differentiable patch attack algorithm and can be readily integrated for adversarial training to defend against high-profile patch attacks.
    \item Experiments show that GDPA has superior attack success rates over strong patch attack baselines, and the adversarially trained model with GDPA is more robust to various adversarial attacks than state-of-the-art methods.
\end{itemize}

\section{Related Works}\label{sec:related}
\vspace{-5pt}
\paragraph{Adversarial Attack}
Most adversarial attack methods focus on adding imperceptible perturbation covering the entire image~\cite{goodfellow2014explaining,szegedy2013intriguing, gao2020patch}. Recently, researchers have shown that perturbing a part of image with perceptible noise is another practical method to attack DNN models~\cite{brown2017adversarial, karmon2018lavan, wu2019defending, sharif2016accessorize, evtimov2017robust, yang2020patchattack, liu2020bias, yang2019design, liu2019perceptual}. Sharif et al.~\cite{sharif2016accessorize} propose to add eyeglasses with a specially constructed frame texture to attack face recognition. Eykholt et al.~\cite{evtimov2017robust} show that adding specific rectangular solid-colored patches on traffic signs can fool traffic sign classification. LAVAN~\cite{karmon2018lavan} learns visible and localized patches that are transferable across images and locations by training the pattern at a random location with a randomly picked image in each iteration. Recently, Wu et al.~\cite{wu2019defending} propose a Rectangle Occlusion Attack (ROA) to generate adversarial patches for adversarial training. ROA uses an exhaustive search (ROA-Exh) or a gradient guided search (ROA-Grad) to find the location that maximizes the cross-entropy (CE) loss and optimizes the patch pattern afterwards. Specifically, ROA-Exh exhaustively searches on images with a stride, and ROA-Grad uses the magnitude of gradient of the CE loss as the sensitivity of regions to identify the top candidate regions to accelerate the location search. However, ROA has some considerable limitations. Firstly, it employs a two-stage attack generation, which separates the process of finding the patch location and patch pattern into two steps: it first finds the position using a \emph{gray pattern} and then optimizes the patch pattern at that position. Hence, the location identified by a \emph{gray pattern} may not be the best patch location for the optimized pattern. Secondly, the two-stage optimization of ROA is computationally expensive and slows down the patch generation process during inference. Different from these algorithms, our GDPA trains a generator to generate the patch pattern and location altogether for each input image. Moreover, GDPA is end to end differentiable, which entails an efficient optimization and easy integration for adversarial training.

Before GDPA, several works~\cite{poursaeed2018generative,baluja2017adversarial,reddy2018nag,xiao2018generating} have proposed to train generators to generate perturbation to improve the fooling rate and inference speed. Poursaeed et al.~\cite{poursaeed2018generative} present a trainable network to transform input images to adversarial perturbations. Baluja and Fischer~\cite{baluja2017adversarial} train feed-forward neural networks in a self-supervised manner to generate adversarial examples against a target network. Different to these generator-based attack methods, our GDPA generates both patch pattern and patch location altogether, and employ an affine transform to synthesize adversarial patch examples.

\vspace{-10pt}
\paragraph{Adversarial Defense}
Defending against adversarial attacks is a challenging task. Different types of defense algorithms have been proposed in the past few years~\cite{dziugaite2016study,guo2017countering,luo2015foveation,gao2017deepcloak,gu2014towards,lyu2015unified,akhtar2018defense,guo2017countering,Xu_2018,wu2019defending, naseer2019local, chiang2020certified, hayes2018visible}, among which adversarial training (AT)~\cite{madry2017towards} has been proved the most effective one against adversarial attacks. AT employs adversarial examples as data augmentation to train a robust model. It has been shown that this method can improve the defense accuracy effectively and sometimes can even improve the accuracy upon the model trained only on the original clean dataset~\cite{wang2020once}. However, a recent work of Wu et al.~\cite{wu2019defending} shows that robust models trained by AT exhibit limited effectiveness against high-profile patch attacks~\cite{sharif2016accessorize,evtimov2017robust,brown2017adversarial}. As the first work attempting to defend patch attacks, Wu et al.~\cite{wu2019defending} propose DOA, which performs a standard adversarial training with Rectangle Occlusion Attack (ROA). As we discussed earlier in this section, ROA has some considerable limitations, which limit its performance on adversarial defense. Our GDPA does not suffer from those limitations of ROA, and is end-to-end differentiable and more amenable for adversarial training.

\section{The GDPA Framework}\label{sec:framework}
\vspace{-10pt}
\begin{figure*}[ht!]
\begin{center}
\includegraphics[width=0.7\linewidth]{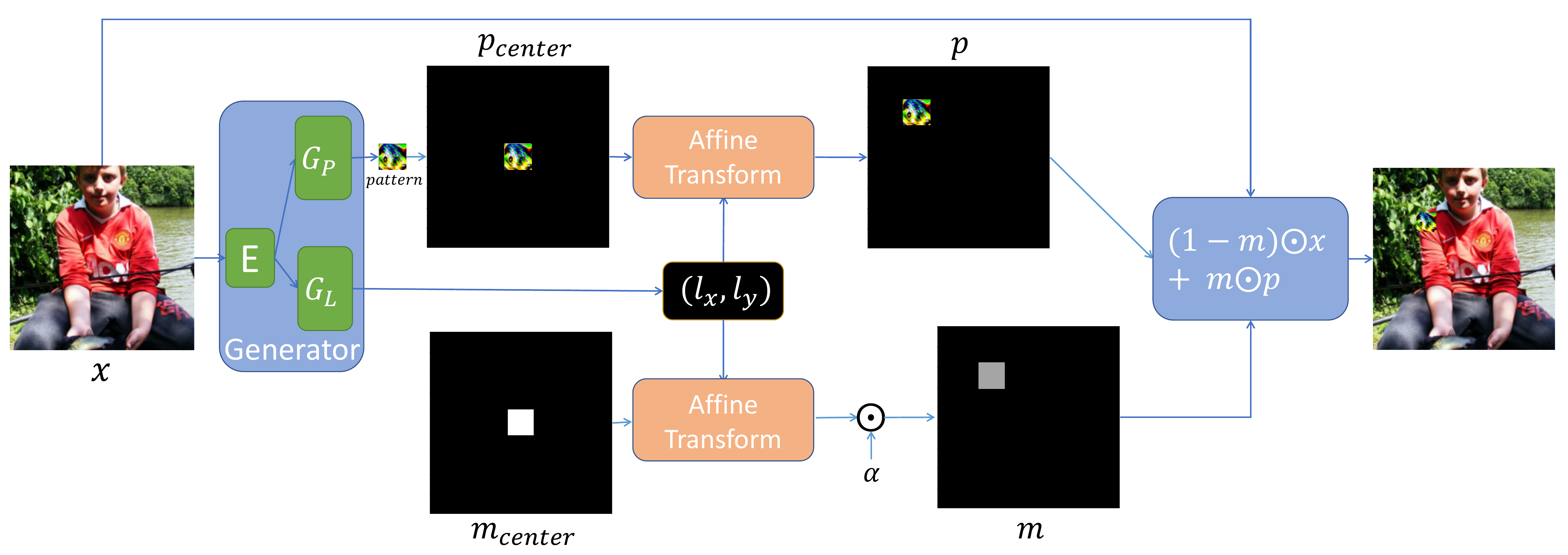}
\end{center}
\vspace{-15pt}
\caption{The GDPA generation pipeline. Given an image $x$, GDPA generates a patch pattern and a patch location for weighted adversarial patch injection. $\alpha\in[0,1]$ controls the visibility of the patch attack. The pipeline is fully differentiable.}
\label{fig:gdpa_arch}
\vspace{-10pt}
\end{figure*}

GDPA is a framework that aims to conduct dynamic patch attack by generating adversarial patch pattern and patch location altogether for each input image. It has a generic formulation that can generate dynamic/static and visible/invisible patch attacks. As an overview, Figure~\ref{fig:gdpa_arch} illustrates the GDPA generation pipeline, while Figure~\ref{fig:at_arch} demonstrates how GDPA can be utilized to train an adversarially robust model.
\vspace{-10pt}
\subsection{Problem Formulation}
We start with the definition of dynamic patch attack. Let $\mathbb{D}=\{\mathcal{X},\mathcal{Y}\}$ denote a training dataset, where $\mathcal{X}$ is a set of images of size $w \times h$, and $\mathcal{Y}$ are their corresponding labels. Let $T:\mathcal{X}\rightarrow\mathcal{Y}$ denote a target model that we attempt to attack. Given an image $x\in\mathcal{X}$ and a target model $T$, our \textit{dynamic patch attack} aims to find a pattern of size $w' \times h'$ and a position in image that once placed on image $x$ it can mislead the target model. 
\vspace{-10pt}
\subsection{Localized Pattern Generation}

One crucial component of GDPA is the generator that generates patch pattern and patch location for a given image. Since patch pattern and patch location are coupled to a given image, we design a generator $G$ with two heads that share the same latent features extracted by an encoder. Specifically, our generator includes an encoder $G_E$ to extract the feature representation of image $x$, followed by a location decoder $G_L$ and a pattern decoder $G_P$ to generate location and pattern of the adversarial patch:
\vspace{-5pt}
\begin{align}
    &l_x, l_y = \tanh{(G_{L}(G_E(x)) / \beta)},\\
    &pattern = 0.5\times\tanh(G_{P}(G_E(x)))+0.5,
\end{align}
where $l_x$ and $l_y$ are the location (2D coordinates) of a patch in image $x$ with the origin at the center of image, and $pattern$ is the patch pattern of size $w' \times h'$. To keep the patch location $l_x$ and $l_y$ within the boundary of image, we use a $\tanh$ function to constrain $l_x$ and $l_y$ in the range of $[-1, 1]$, where $\beta$ is a hyperparameter that controls the slope of $\tanh$. All experiments in this paper use $\beta=3000$, which we found to work well across a variety of architectures and datasets. Similarly, we use another $\tanh$ to impose the pattern values in the range of $[0,1]$\footnote{As a preprocessing step, all images are normalized to have pixel values in the range of $[0,1]$.}.

Specially, we use a convolutional neural network as our encoder network $G_E$, with an architecture adapted from the work of image-to-image translation~\cite{CycleGAN2017}. On top of $G_E$, we use two fully-connected networks as our decoders $G_P$ and $G_L$, respectively. Due to page limit, details of the network architectures are provided in the Appendix.

\subsection{Weighted Adversarial Patch Injection}

With the generated patch location and patch pattern, we then define a function to inject the patch into image $x$. Standard adversarial attacks~\cite{madry2017towards} employ an additive function to inject noise: $x^\prime = x + p$, where $p$ is an imperceptible adversarial perturbation. Recently, other forms of perturbations, such as multiplicative ones $x^\prime = x\odot m$~\cite{xvat2020}, have been explored to inject perturbations. In addition, LAVAN~\cite{karmon2018lavan} employs $(1-m) \odot x + m \odot p $ with a binary mask $m \in \{0, 1\}^{w'\times h'}$ to generate patch attack adversarial examples. Inspired by LAVAN, we extend this function by relaxing the binary mask to a continuous mask $m \in [0, 1]^{w'\times h'}$ for adversarial patch injection. Specifically, we employ the weighted adversarial patch injection $x^{adv} = (1-m) \odot x + m \odot p$, with $m\in[0, 1]^{w'\times h'},$
which is a convex combination of original image $x$ and patch pattern $p$ with the weight defined by $m$. We find this relaxed version is more flexible and easier to optimize than the one LAVAN explored. Next, we discuss how to use the generated $(l_x, l_y)$ and $pattern$ to inject an adversarial patch to image $x$.
\vspace{-10pt}
\subsection{Differentiable Affine Transformation}

We employ an affine transformation in GDPA to inject adversarial patches into images. To make the whole pipeline differentiable w.r.t. $l_x$ and $l_y$, a bilinear interpolation is used to estimate the pixel values that are not on the pixel grids after transformation. By doing this, the whole pipeline is fully differentiable and the gradient can be back-propagated end-to-end to update parameters of generator $G$. Specifically, we adopt the affine transformation and image sampling method of Spatial Transformer Networks~\cite{jaderberg2015spatial} to define a differentiable translate operator, which can translate a source image to a target image by a displacement of $(l_x, l_y)$. 

We first use an affine transform to compute the pixel index relationship between source image and target image: \vspace{-15pt}
{\scriptsize
\begin{align}
\left( {
\renewcommand{\arraystretch}{1.2}
\begin{array}{c}
x_i^s\\
y_i^s
\end{array}
}\right)
=
\left[ {
\renewcommand{\arraystretch}{1.2}
\begin{array}{ccc}
\theta_{11} & \theta_{12} & \theta_{13}\\
\theta_{21} & \theta_{22} & \theta_{23}
\end{array}
}\right]
\left( {
\renewcommand{\arraystretch}{1.2}
\begin{array}{c}
x^t_i\\
y^t_i\\
1
\end{array}
}\right),
\end{align}
}%
where $(x_i^t, y_i^t)$ is the pixel index of target image, and $(x_i^s, y_i^s)$ is the corresponding pixel index in source image. We set $\theta_{11} = 1, \theta_{21} = 0, \theta_{12} = 0, \theta_{22} = 1, \theta_{13} = w' / 2 \cdot l_x$ and $\theta_{23} = h' / 2 \cdot l_y$ for translation purpose\footnote{Note that we can also learn $\theta_{11}, \theta_{12}, \theta_{21}, \theta_{22}$ to rotate, dilate or shear an adversarial patch to further improve GDPA's performance. For simplicity and also because we can already achieve state-of-the-art attack success rate (ASR) with translation, in this paper we only consider translation and will leave advanced affine transform to future work.}. Thus, we have $x_i^s = x_i^t + w' / 2 \cdot l_x$ and $y_i^s = y_i^t + h' / 2 \cdot l_y$, where $l_x,l_y\in[-1,1]$. Since $(x_i^s, y_i^s)$ are continuous variables, we can use a bilinear interpolation to sample the pixel values from source image:
\vspace{-5pt}
{\small
\begin{align}
v_i\!=\!\!\!\sum_{j=0}^{w\!-\!1} \!\sum_{k=0}^{h\!-\!1}\!\!u_{jk}\max(0,\! 1\!-\!|x_i^s\!\!-\!j|)\max(0,\!1\!-\!|y_i^s\!\!-\!k|),
\label{eqn:bilinear}
\end{align}
}%
where $u_{jk}$ is the pixel value at index $(j,k)$ of the source image, and $v_i$ is the output value of pixel $i$ at index $(x_i^t, y_i^t)$ of the translated image. With the affine transform and bilinear sampler described above, we have a differentiable translate operator, which we denote as $Translate()$ in the rest part of the paper.

\begin{figure*}[ht!]
\begin{center}
\includegraphics[width=0.7\linewidth]{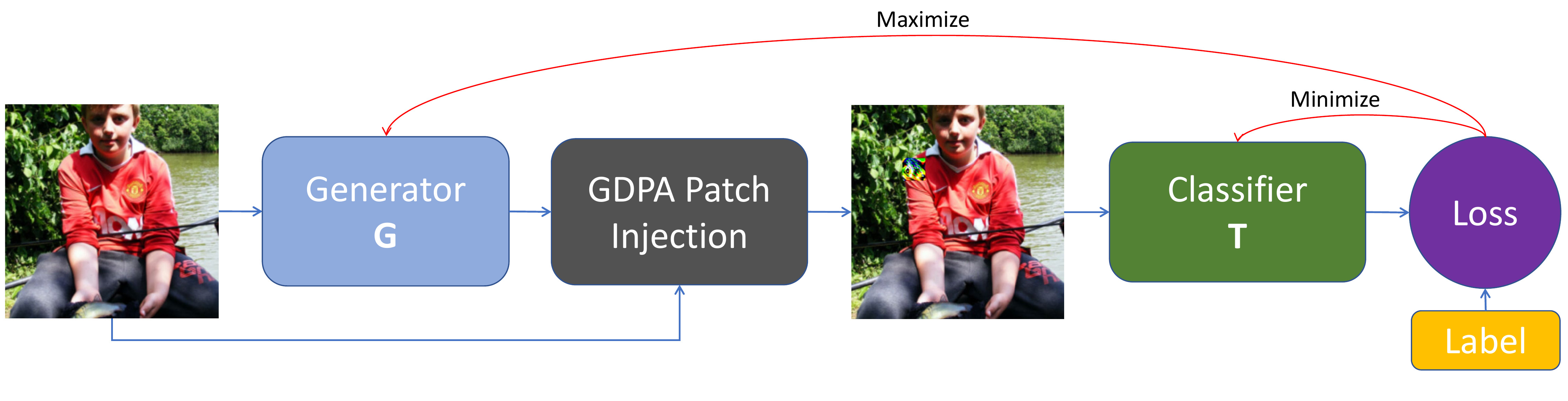}
\end{center}
\vspace{-20pt}
\caption{The GDPA-AT pipeline. Given an image, GDPA generates an adversarial patch to maximize the loss of classifier $T$, while classifier $T$ learns from the patch attack to minimize its loss.}
\label{fig:at_arch}
\end{figure*}
\vspace{-10pt}
\subsection{Generative Dynamic Patch Attack}
Figure~\ref{fig:gdpa_arch} illustrates the GDPA generation pipeline, which includes the three components we described above: patch pattern and location generator, differentiable affine transform, and the weighted adversarial patch injection to produce an image-dependent dynamic patch attack.

As shown in Figure~\ref{fig:gdpa_arch}, we introduce an initial mask $m_{center}$ of the same size of input image, and the center part of the mask has value 1 and rest of 0. Then we use the affine transform $Translate()$ to translate $m_{center}$ by a displacement of $(l_x, l_y)$: \vspace{-5pt}
\begin{align}\label{eq:visible_m}
  m = \alpha \cdot Translate(m_{center}, l_x, l_y),
\end{align}
where $\alpha\in[0,1]$ is a hyperparameter that controls the visibility of adversarial patches. When $\alpha=1$, the patch would be completely visible and replace the original image pixel values; otherwise, the visibility of the adversarial patch will be lower. In practice, we can use a small value of $\alpha$ to generate human imperceptible adversarial patches.

Similarly, we can generate a translated patch pattern. As shown in Figure~\ref{fig:gdpa_arch}, once $pattern$ is generated, we zero-pad it to create a pattern $p_{center}$ of the same size of input image with $pattern$ at the center. We then translate $p_{center}$ by $(l_x, l_y)$ via the affine transform: $p = Translate(p_{center}, l_x, l_y).$


Finally, we can generate a GDPA adversarial example for image $x$ by $x^{adv} = (1-m) \odot x + m \odot p.$
As we can see, all the components in Figure~\ref{fig:gdpa_arch} are differentiable. Therefore, the whole GPDA generation pipeline is fully differentiable and can be optimized efficiently with gradient-based methods.

\vspace{-25pt}
\begin{align}
  \argmin_G -\Lagr_{\texttt{CE}}(T, x^{adv}, y). 
\end{align}
We can also launch a targeted patch attack to fool the target model $T$ to misclassify an input $x$ as target class
\vspace{-10pt}
\begin{align}
  \argmin_G \Lagr_{\texttt{CE}}(T, x^{adv}, y_{target}).
\end{align} 
Details of the GDPA training algorithm can be found in Algorithm~\ref{alg:GDPA}.
\vspace{-10pt}

\subsection{Adversarial Training with GDPA}
Adversarial training with the PGD attack exhibits limited effectiveness against high-profile patch attacks~\cite{wu2019defending}. In this section, we discuss how to utilize GDPA for adversarial training to improve model robustness against high-profile patch attacks.

Figure~\ref{fig:at_arch} illustrates the GDPA adversarial training (GDPA-AT) pipeline to train a robust model against patch attacks. Similar to Generative Adversarial Networks~\cite{gan}, GDPA-AT trains generator $G$ and target classifier $T$ iteratively to optimize the following minimax objective: \vspace{-10pt}
{\small
\begin{align}
  \min_T\max_G\mathbb{E}_{(x,y)\sim\mathbb{D}}[\Lagr_{\texttt{CE}}(T, x^{adv}, y)],
\end{align}
}%
where the inner maximization step optimizes generator $G$ to maximize the classification loss of $T$, while the outer minimization step optimizes target classifier $T$ to minimize the classification loss. Unlike the traditional adversarial training, in which the inner maximization step usually optimizes an adversarial example $x^{adv}$ directly, our GDPA-AT optimizes a generator $G$ to generate patch attack with one forward propagation. As the iterative training proceeds, the generator $G$ searches for the weakest image region to attack classifier $T$ at each iteration, while $T$ learns from the current patch attacks and becomes more resilient to these attacks over time. Details of our GDPA-AT algorithm are described in 
Algorithm~\ref{alg:at}.

\begin{minipage}[t]{6.2cm}
\vspace{0pt}  
\begin{algorithm}[H]
  \footnotesize
  \SetAlgoLined
  \KwIn{training set $\mathbb{D}$; target model $T$; visibility $\alpha$}
  \KwOut{generator $G$}
    initialize generator $G$\;
    \For(){number of training epochs}
    {\For(){each $(x, y)\in\mathbb{D}$}
    {
     $l_x, l_y = G_L((G_E(x))$\; 
     $pattern = G_P(G_E(x))$\; 
     $m = \alpha\cdot Translate(m_{center}, l_x, l_y)$\; 
     $p = Translate(p_{center}, l_x, l_y)$\; 
     $x^{adv} = (1-m) \odot x + m \odot p$\;
     \eIf{targeted attack}{
     $loss = \Lagr_{\texttt{CE}}(T, x^{adv}, y_{target})$\;
     }{
     $loss = -\Lagr_{\texttt{CE}}(T, x^{adv}, y)$\;
     }
     $\theta_G = \theta_G - lr * \partial loss / \partial \theta_G $
    }
    }
  \caption{GDPA Training}\label{alg:GDPA}
\end{algorithm}
\small
\end{minipage}
\begin{minipage}[t]{6.2cm}
\vspace{0pt}  
\begin{algorithm}[H]
    \footnotesize
    \SetAlgoLined
    \KwIn{training set $\mathbb{D}$}
    \KwOut{target classifier $T$; generator $G$}
     initialize classifier $T$ and generator $G$\;
     \For(){number of training epochs}
     {\For(){each $(x,y)\in\mathbb{D}$}
     {
     $x^{adv} = GDPA(G, x)$ \;
     $loss = -\Lagr_{\texttt{CE}}(T, x^{adv}, y)$ \;
     $\theta_G = \theta_G - lr_G * \partial loss / \partial \theta_G $ \;
     $x^{adv} = GDPA(G, x)$ \;
     $loss = \Lagr_{\texttt{CE}}(T, x^{adv}, y)$ \;
     $\theta_T = \theta_T - lr_T * \partial loss / \partial \theta_T $
     }}
     \caption{GDPA-AT Training}
     \label{alg:at}
\end{algorithm}
\end{minipage}

\vspace{-10pt}
\section{Experimental Results}

We now validate GDPA on benchmark datasets for adversarial patch attack and adversarial defense. Specifically, we evaluate the performance of GDPA on patch attack in Section~\ref{sec:exp_attack} and GDPA-AT on improving model robustness in Section~\ref{sec:exp_at}. To evaluate the inference efficiency, we also compare the run-times of GPDA and state-of-the-art attack algorithms in Section~\ref{sec:exp_speed}. All our experiments are performed with PyTorch on Nvidia RTX GPUs. Our source code is provided as a part of supplementary materials.

\vspace{-10pt}
\paragraph{Experimental Setup}

We evaluate GDPA and GDPA-AT on three benchmark datasets: VGGFace~\cite{sharif2016accessorize}, Traffic Sign~\cite{evtimov2017robust} and ImageNet~\cite{deng2009imagenet}. To evaluate GDPA's attack performance, we compare GDPA with LAVAN~\cite{karmon2018lavan} and ROA~\cite{wu2019defending}, two state-of-the-art patch attack algorithms that generate patches based on iterative optimizations. Following their experimental settings, we run LAVAN and ROA for 50 optimization iterations with a learning rate of 4. For adversarial defense experiments, we compare GDPA-AT with DOA~\cite{wu2019defending} and PGD-AT~\cite{madry2017towards}. The former is a state-of-the-art defense algorithm for patch attacks, while the latter is a well-established defense algorithm for adversarial attacks. We evaluate the robustness of the models under eyeglasses attack~\cite{sharif2016accessorize}, sticker attack~\cite{evtimov2017robust}. Following the settings in DOA~\cite{wu2019defending}, we use $70\times70$ patches with stride 5 for VGGFace and $7\times7$ patches with stride 2 for Traffic Sign to generate ROA attacks. We set $\epsilon\!=\!16$ for PGD-AT since this yields the best results of PGD-AT. We use attack success rate (ASR)~\cite{dong2020benchmarking} as the metric to evaluate the effectiveness of an attack, and use classification accuracy to evaluate the robustness of a model when under adversarial attacks. Details of benchmark datasets, high-profile patch attacks, network architectures and training procedures can be found in the Appendix. 
\vspace{-10pt}
\subsection{Dynamic Patch Attack} \label{sec:exp_attack}

We first evaluate the performance of GDPA on non-targeted and targeted patch attacks and compare it with the state-of-the-arts: LAVAN~\cite{karmon2018lavan} and ROA~\cite{wu2019defending}. We provide results of two versions of ROA: ROA-Exh and ROA-Grad, where the former exhaustively searches for a patch location in images with a fixed stride, and the latter uses the magnitude of gradient as the sensitivity of regions to identify top regions to accelerate the location search. We evaluate the effectiveness of the attack algorithms when perturbing different percentages of pixels. To interpret the results, we also visualize the perturbed images generated by GDPA.

\begin{table}[h!]
\begin{center}
\resizebox{0.8\columnwidth}{!}{
\begin{tabular}{l|l|cccccccc}
  \toprule
  &  & \multicolumn{8}{c}{Percentage of Attacked Pixels} \\
  \cmidrule(lr){3-10}
  \multicolumn{1}{c|}{Dataset} & \multicolumn{1}{c|}{Algorithm} & \multicolumn{4}{c}{Non-Targeted Attack} & \multicolumn{4}{c}{Targeted Attack} \\ 
  \cmidrule(lr){3-6}\cmidrule(lr){7-10}
  & & 1\%     & 2\%     & 5\%     & 10\%    &  1\%     & 2\%     & 5\%     & 10\%     \\ 
  \midrule
    \multirow{3}{*}{Traffic Sign}& LAVAN~\cite{karmon2018lavan} & 33.4  & 58.7  & 85.1  & 93.9  & 32.2  & 48.1  & \textbf{80.9}  & 89.9 \\
  & ROA-Grad~\cite{wu2019defending}  & 36.2  & 61.8  & 87.3  & 93.6 & 29.8  & 44.6  & 74.5  & 90.5\\ 
  & ROA-Exh~\cite{wu2019defending}  & 37.1   & 63.0   & 89.4   & 93.8  & 31.3   & 45.9   & 76.2  & 91.7\\ 
  & \textbf{GDPA}  & \textbf{39.6}  & \textbf{64.1}  & \textbf{91.3}  & \textbf{94.3} & \textbf{33.9}  & \textbf{50.4}  & 77.5  & \textbf{92.8} \\ 
  \midrule
  \multirow{3}{*}{VGGFace}& LAVAN~\cite{karmon2018lavan} & 31.9  & 42.7  & 56.3  & 92.0  & 37.8  & 57.9  & 67.2  & 94.6  \\
  & ROA-Grad~\cite{wu2019defending}  & 37.5  & 62.3  & 84.2  & \textbf{99.6} & 46.3  & 75.6  & 89.0  & 99.2  \\ 
  & ROA-Exh~\cite{wu2019defending}  & 38.3  & 64.5   & 86.0   & \textbf{99.6}  & 48.2  & 76.7   & 91.1   & 99.3 \\ 
  & \textbf{GDPA}  & \textbf{46.3}  & \textbf{76.4}  & \textbf{88.4}  & 99.5 & \textbf{50.5}  & \textbf{83.4}  & \textbf{95.5}  & \textbf{99.8}  \\ 
  \midrule
  \multirow{3}{*}{ImageNet} & LAVAN~\cite{karmon2018lavan} & 89.2  & 92.8  & 97.8  & \textbf{99.9} & 86.3  & 93.8  & \textbf{99.7}  & 99.8   \\ 
  & ROA-Grad~\cite{wu2019defending}  & 93.5  & 94.6  & 98.7  & 99.7 & 79.6  & 88.3  & 97.5  &   99.8 \\ 
  & ROA-Exh~\cite{wu2019defending}  &  94.8 & 95.3  & 99.2  & 99.7 & 81.1  & 89.6   & 98.4   & 99.8 \\ 
  & \textbf{GDPA} & \textbf{96.3}  & \textbf{96.9}  & \textbf{99.7}  & 99.8 & \textbf{89.3}  & \textbf{94.4}  & 99.6  & \textbf{99.9}    \\ 
  \bottomrule
\end{tabular}
}
\end{center}
\caption{The ASRs of different patch attack algorithms on datasets Traffic Sign, VGGFace and ImageNet. Both non-targeted attack and targeted attack are considered. The performances are evaluated with patches of different sizes.}\label{tab:asr_size}
\vspace{-10pt}
\end{table}

Table~\ref{tab:asr_size} reports the ASRs of GDPA and the other competing algorithms for non-targeted and targeted patch attacks. The ASRs of an attack algorithm are evaluated on a model trained with cross-entropy (CE) loss when attacked with patches of different sizes (1\%, 2\%, 5\% or 10\% of pixels). Specifically, We use square patches of width 3, 5, 7, 10 for Traffic Sign and 23, 32, 50, 71 for VGGFace and ImageNet. For targeted attacks, we choose the first class for each of the three datasets as the target class, i.e., ``AddedLine", ``Aamir Khan" and ``tench, Tinca tinca", respectively. As expected, the larger patch size is, the higher ASR is achieved for all patch attack algorithms. In most of the cases, GDPA achieves higher ASRs than the competing algorithms.

\begin{figure}[h!]
\centering
\includegraphics[width=0.9\linewidth]{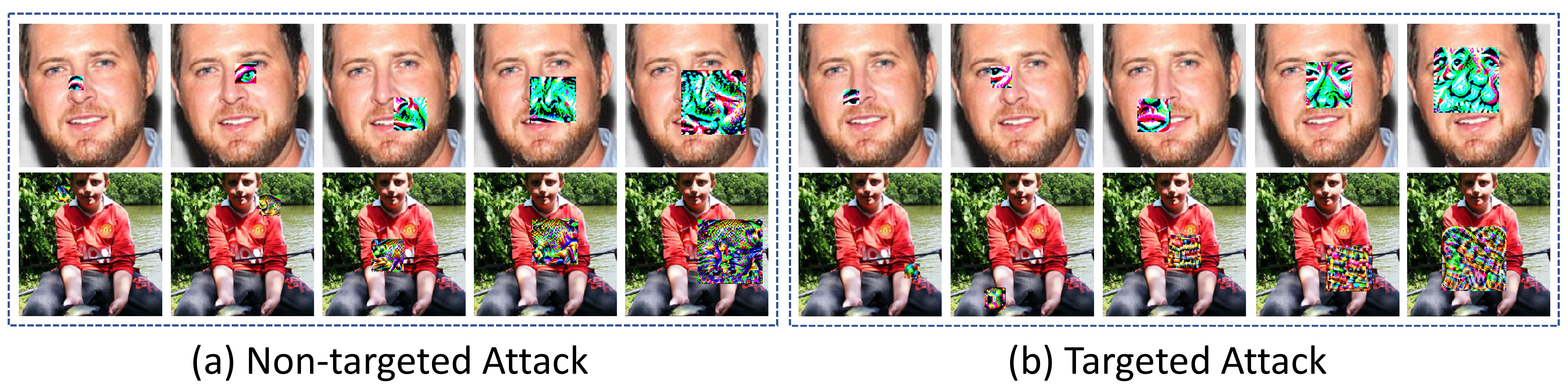}
\vspace{-10pt}
\caption{Perturbed images of VGGFace and ImageNet generated by GDPA with different patch sizes. The last column of targeted attack are example images of target classes.}
\label{fig:attack_vis}
\end{figure}

Figure~\ref{fig:attack_vis} visualizes the perturbed images of different patch sizes generated by GDPA for non-targeted and targeted attack on VGGFace and ImageNet. As we can see, the patches generated on VGGFace (top row) demonstrates clear semantic meanings, resembling human eyes, nose or mouth. Moreover, the positions chosen by GDPA on face images are in a close proximity of the original face features. On the other hand, the patches generated on ImageNet (bottom row) do not demonstrate a strong semantic meaning with non-targeted attack but shows semantic meaning with targeted attack. 





\begin{figure}
\begin{floatrow}
\ffigbox[\FBwidth]{
  \includegraphics[width=0.8\linewidth]{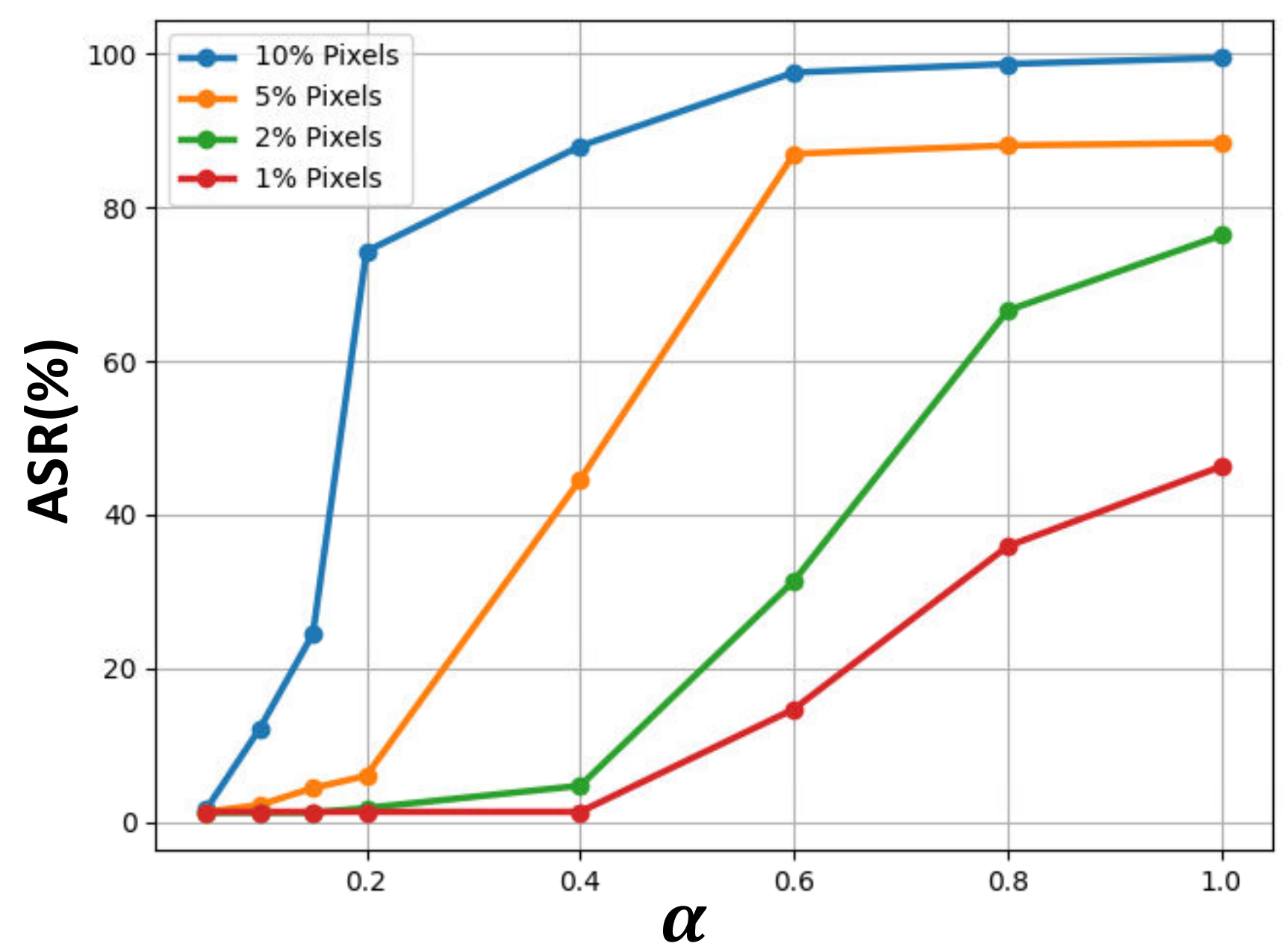}
}{
  \caption{The impact of $\alpha$ to the ASR of GDPA on VGGFace.}\label{fig:asr_alpha}
}
\capbtabbox{
{\scriptsize
\setlength\tabcolsep{2.5pt}%
\begin{tabular}{l|l|cccccccc}
  \toprule
  &  & \multicolumn{8}{c}{Percentage of Attacked Pixels} \\
  \cmidrule(lr){3-10}
  \multicolumn{1}{c|}{Dataset} & \multicolumn{1}{c|}{\begin{tabular}[c]{@{}l@{}}Patch\\ Location\end{tabular}} & \multicolumn{4}{c}{Non-Targeted Attack} & \multicolumn{4}{c}{Targeted Attack} \\ 
  \cmidrule(lr){3-6}\cmidrule(lr){7-10}
  & & 1\%     & 2\%     & 5\%     & 10\%    &  1\%     & 2\%     & 5\%     & 10\%     \\ 
  \midrule
    \multirow{3}{*}{Traffic Sign}& 
   Random  & 35.1   & 59.4   & 86.3   & 93.5  & 28.9   &  44.1  & 75.6  & 90.1\\ 
  & Dynamic  & \textbf{39.6}  & \textbf{64.1}  & \textbf{91.3}  & \textbf{94.3} & \textbf{33.9}  & \textbf{50.4}  & \textbf{77.5}  & \textbf{92.8} \\ 
  \midrule
  \multirow{3}{*}{VGGFace}& Random & 35.4  & 58.8  & 82.7  & 96.5  & 43.9  &  71.7 & 88.3  & 97.9  \\
  
  & Dynamic  & \textbf{46.3}  & \textbf{76.4}  & \textbf{88.4}  & \textbf{99.5} & \textbf{50.5}  & \textbf{83.4}  & \textbf{95.5}  & \textbf{99.8}  \\ 
  \midrule
  \multirow{3}{*}{ImageNet} & Random & 91.6  & 93.5  & 98.1  & 99.7 & 79.3  & 87.0  & 96.6  &  99.8  \\ 
  & Dynamic & \textbf{96.3}  & \textbf{96.9}  & \textbf{99.7}  & \textbf{99.8} & \textbf{89.3}  & \textbf{94.4}  & \textbf{99.6}  & \textbf{99.9}    \\ 
  \bottomrule
\end{tabular}
  }
}{
  \caption{The ASRs of GDPA with patches of dynamic or random locations on datasets Traffic Sign, VGGFace and ImageNet.}\label{tab:gdpa_rand}
}
\end{floatrow}
\vspace{-10pt}
\end{figure}

\vspace{-10pt}
\paragraph{Visibility $\alpha$ vs. ASR}
We further investigate the impact of visibility parameter $\alpha$ of Eq.~\ref{eq:visible_m} to GDPA's ASR. The results on VGGFace are shown in Figure~\ref{fig:asr_alpha}, where we consider different patch sizes. As expected, when $\alpha$ increases, the attack strength of GDPA gets stronger for all different patch sizes. Notably, when the patch size is 5\% or 10\% of pixels, GDPA can reach almost the highest ASRs when $\alpha\geq 0.6$, indicating that when patches are sufficiently large, the attack can be more invisible to attack a model successfully. Example perturbed images generated by GDPA with different $\alpha$ are provided in the Appendix.

\paragraph{Dynamic vs. Random Patch Location} Tab~\ref{tab:gdpa_rand} shows the comparison of GDPA framework with random patch locations or dynamic patch locations. As we can see from the table, GDPA achieves higher ASR with dynamic patch location than random location. This shows that learned image-dependant dynamic locations contribute to the superior performance of GDPA.

\begin{table}[h!]
\begin{center}
\resizebox{0.95\columnwidth}{!}{
  \begin{tabular}{lcccccccccccc}
  \toprule
  & \multicolumn{4}{c}{Eyeglasses Attack (VGGFace)} & \multicolumn{4}{c}{Sticker Attack (Traffic Sign)} & \multicolumn{4}{c}{LAVAN (ImageNet)} \\
  \cmidrule(lr){2-5}\cmidrule(lr){6-9}\cmidrule(lr){10-13}
  Attack Strength & 0 & 100 & 200 & 300 & 0 & 10 & 100 & 1000 & 0\% & 1\% & 2\% & 5\%   \\ 
  \midrule
  CE Training          & 98.9  & 0.2   & 0.1   & 0.0  & \textbf{98.7}  & 42.9  & 32.5  & 24.3 & \textbf{76.6} & 41.4 & 0.7 & 0.1  \\ 
  PGD-AT~\cite{madry2017towards} & 97.3  & 37.7  & 36.9  & 36.6 & 97.5  & 57.6  & 45.1  & 42.5 & 62.4 & 45.4 & 37.8 & 12.0    \\
  \midrule
  DOA-Grad~\cite{wu2019defending}             & 99.2  & 85.3  & 83.7  & 81.9  & 95.6  & 85.2  & 84.8  & 82.8 & 61.9 & 52.8 & 51.1 & 47.3   \\ 
  DOA-Exh~\cite{wu2019defending}               & 99.0  & 89.8  & 87.8  & 85.9 & 92.9  & 92.2  & 91.8  & 90.8 & 62.0 & 54.2 & 53.7 & 50.6   \\ 
  \textbf{GDPA-AT}                 & \textbf{99.6}  & \textbf{96.5}  & \textbf{94.7}  &  \textbf{94.9}      & 98.5  & \textbf{96.2}  & \textbf{95.8}  & \textbf{94.6}  & 62.2 & \textbf{58.3} & \textbf{56.9} & \textbf{53.2} \\ \bottomrule
  \end{tabular} 
  }
  \end{center}
  \caption{The accuracies of different robust models under eyeglasses attack on VGGFace (left), sticker attack on Traffic Sign  (middle), and LAVAN on ImageNet (right). We evaluate under different iterations for Eyeglasses Attack and Sticker Attack , and different patch size for LAVAN.}\label{tab:at}
  \vspace{-10pt}
\end{table}

\vspace{-10pt}
\subsection{Dynamic Patch Adversarial Training} \label{sec:exp_at}

Next we validate the robustness of models trained by GDPA-AT against various adversarial patch attacks. Specifically, we report the results of GDPA-AT trained models against Eyeglass Attack, Sticker Attack and LAVAN, and compare them with state-of-the-art defense methods.

Table~\ref{tab:at} reports the accuracies of robust models trained by different defense algorithms against three types of patch attacks: 1) eyeglasses attack on VGGFace, 2) sticker attack on Traffic Sign and 3) LAVAN on ImageNet. As can be seen, PGD-AT, a well-established defense method for conventional adversarial attacks, is not robust to all three patch attacks, which is consistent with the results reported in~\cite{wu2019defending}. While both DOA and GDPA-AT improve the robustness over PGD-AT significantly, GDPA-AT achieves substantially higher accuracies than the two variants of DOA. 

\vspace{-10pt}
\subsection{Inference Speed}\label{sec:exp_speed}
\vspace{-5pt}
Besides the improved attack and defense performance of GDPA, another advantage of GDPA is its superior inference speed to generate attacks over the optimization-based methods, such as PGD~\cite{madry2017towards} and ROA~\cite{wu2019defending}. To have a quantitative comparison in terms of inference time, we evaluate the run-time of GDPA, PGD and ROA on the VGGFace test dataset (470 images). GDPA needs one forward propagation to generate a patch attack, while we follow the settings of ROA and PGD and run 50 iterative optimizations to generate their attacks. As shown in Table~\ref{tab:speed}, GDPA is about 40x faster than PGD and 47x faster than ROA. 

\begin{table}[ht!]
\begin{center}
{\small
\begin{tabular}{lcccc}
\toprule
& PGD~\cite{madry2017towards} & ROA-Grad~\cite{wu2019defending} & ROA-Ex~\cite{wu2019defending} & \textbf{GDPA} \\ 
\midrule
Inference-time (s) & 108.31 & 129.42 & 248.82 & \textbf{2.76}         \\ 
\bottomrule
\end{tabular}
}
\end{center}
\vspace{-10pt}
\caption{Inference-time comparison of different attack algorithms on the VGGFace test dataset (470 images).}
\label{tab:speed}
\end{table}

\vspace{-20pt}
\paragraph{Additional Experimental Results}
As GDPA is a generic attack algorithm, we conduct additional experiments to evaluate its performance with different configurations and also validate some design choices. Due to page limit, details are relegated to the Appendix.

\vspace{-15pt}
\section{Conclusion}
\vspace{-5pt}
This paper introduces GDPA, a novel \emph{dynamic} patch attack algorithm, that generates patch pattern and patch location altogether for each input image. Due to its generic formulation, GDPA can generate dynamic/static and visible/invisible patch attacks. GDPA is end-to-end differentiable, which entails an efficient optimization and easy integration for adversarial training. We validated our method on multiple benchmarks with different model architectures. GDPA demonstrates superior ASR over strong patch attack methods, and the adversarially trained model with GDPA is more robust to high-profile patch attacks. Moreover, GDPA is 40-50x faster than competing attack algorithms, making it a highly effective attack and defense algorithm.

\vspace{-15pt}
\section{Acknowledgment}
We would like to thank the anonymous reviewers for their comments and suggestions, which helped improve the quality of this paper. We would also gratefully acknowledge the support of Cisco Systems Inc. for its research fund to this work.

\bibliography{egbib}

\clearpage
\appendix
\newpage

\section{Experimental Details}

We first describe the three benchmark datasets and target models used in our experiments. These datasets are used to train our GDPA generator, robust models with adversarial training, and evaluate the performance of patch attacks.

\subsection{VGGFace}
\paragraph{Dataset}
The VGGFace dataset~\cite{parkhi2015deep} is a benchmark for face recognition, containing 2,622 subjects and 2.6 million images in total. Same with DOA~\cite{wu2019defending}, we choose 10 subjects and sample face images only containing those individuals. We process the data to the size of $224\times 224$ by standard crop-and-resize, and perform class-balanced split to generate training, validation, and test datasets with ratio 7:2:1. As a result, we obtain 3178, 922 and 470 images for training, validation and test, respectively. The training set is used to train the target model, the GDPA generator and robust models with adversarial training. Likewise, the test set is used to evaluate the target model, the performance of patch attack and adversarial defense. 

\paragraph{Target Model}
We use the VGGFace CNN model~\cite{parkhi2015deep} as the target classifier in our experiments. We use standard transfer learning on our processed dataset, keeping the convolutional layers in the VGGFace CNN model, but adjusting the number of output neurons of the last fully connected layer to 10. In order to use the pre-trained weights from the convolutional layers of VGGFace CNN model, we convert the images from RGB to BGR and subtract the mean value $[129.2, 104.8, 93.6]$. We set the batch size to 64 and use the Adam Optimizer with an initial learning rate of $10^{-4}$. We drop the learning rate by 0.1 every 10 epochs. For hyperparameter tuning and model selection, we track the accuracy on validation set to avoid overfitting. We train the model on training set for 30 epochs and obtain an accuracy of 98.94\% on test data.

\subsection{Traffic Sign}
\paragraph{Dataset}
To have a fair comparison with DOA~\cite{wu2019defending}, we pick the same 16 traffic signs from the dataset LISA~\cite{mogelmose2012vision} with 3,509 training and 1,148 validation images. Following the prior works~\cite{evtimov2017robust,wu2019defending}, we further sample 40 stop signs from the validation set as the test data to evaluate performance of the stop sign classification. Similarly, all the data are processed by standard crop-and-resize to $32\times 32$ pixels. Same with VGGFace, we use the training set to train the target model, the GDPA generator and robust models with adversarial training. We use the test set to evaluate the performance of the target model, patch attack and adversarial defense.

\paragraph{Target Model}
We use the LISA-CNN~\cite{evtimov2017robust} as the target model, which contains three convolutional layers and one fully-connected layer. We use the Adam Optimizer with initial learning rate 0.1 and drop the learning rate by 0.1 every 10 epochs. We set the batch size to 128. After 30 epochs, we achieve an accuracy of 98.69\% on the validation set, and 100\% accuracy on the test data.

\subsection{ImageNet}
\paragraph{Dataset}
ImageNet~\cite{deng2009imagenet} is a well-known large scale object recognition benchmark. To develop the training and validation sets to train and evaluate the GDPA generator and robust models with adversarial training, we follow Moosavi-Dezfooli et al.~\cite{moosavi2017universal} to select a subset of $10,000$ images from ImageNet training set (randomly choose ten images for each class) as our training set, and use the whole ImageNet validation set ($50,000$ images) as our validation set.

\paragraph{Target Model}
Following Poursaeed at el.~\cite{poursaeed2018generative}, we use a pre-trained VGG19 model~\cite{simonyan2014very} from PyTorch library as the target model. This model achieves an accuracy of 72.4\% on the validation set.

\section{Patch Attacks}

\paragraph{Eyeglasses Attack}
This is an effective physically realizable patch attack developed by Sharif et al.~\cite{sharif2016accessorize}. It first initializes the eyeglass frames with 5 different colors, and chooses the color with the highest cross-entropy loss as starting color. For each update step, it divides the gradient value by its maximum and multiplies the results with the learning rate. Then it only keeps the gradient value in the eyeglass frame area. Finally, it clips and rounds the pixel values to keep them in the valid range. We evaluate the eyeglasses attack on the test set of VGGFace.

\paragraph{Sticker Attack}
Proposed by Evtimov et al.~\cite{evtimov2017robust}, this is another physically realizable patch attack. It initializes the stickers on the stop signs with random noise at fixed locations. For each update step, it uses the Adam optimizer with the learning rate 0.1 (and default parameters) to maximize the classification loss of the target model. Just as the other patch attacks, adversarial perturbations are restricted to the mask area; in our experiments, we use the same collection of small rectangles as in~\cite{evtimov2017robust}. We evaluate the sticker attack on the test set of Traffic Sign.

\section{GDPA Network Architecture and Training Details}

\paragraph{Network Architecture}
For VGGFace and ImageNet, both having images of size $224\times224$, we adopt the encoder network structure $G_E$ from the work of image-to-image translation~\cite{CycleGAN2017}. For the Traffic Sign dataset, which has images of size $32\times 32$, we adopt a CNN of 3 convolutional layers with kernel size 4 and stride 2 as the encoder network $G_E$. We then use a neural network of one fully-connected layer with output size $3 \times w' \times h'$ as the pattern decoder $G_P$, and a neural network of one fully-connected layer with output size $2$ as the location decoder $G_L$.

\paragraph{GDPA Training Details}
Following Algorithm 1, we train the GDPA generator $G$ by using the Adam optimizer with an initial learning rate of 0.1 for VGGFace and ImageNet, and 0.01 for Traffic Sign. We drop the learning rate by 0.2 every 10 epochs and train the generator for 30 epochs. We set the batch size to 32 and $\beta$ to 3000, which we find works well across various architectures and datasets in our experiments.

\paragraph{GDPA-AT Training Details}
Following Algorithm 2, we train the GDPA generator $G$ and target model $T$ iteratively. We initialize the generator with a pre-trained GDPA generator and the target model with a cross-entropy trained model. We set the $w'$ and $h'$ to 70 for VGGFace and Imagenet and 7 for Traffic Sign during the adversarial training. We use the Adam optimizer to train the generator and the target model, with a learning rate of 0.0001 for both VGGFace and Traffic Sign and 0.001 for imagenet, and drop the learning rate by 0.2 every 50 epochs. We use batch size 32 and train for 1000 epochs for VGGFace, 100 epochs for Imagenet and 5000 epochs for Traffic Sign.

\section{Ablation Study}

\subsection{Generate $pattern$ vs $p$}
Instead of generating $pattern$ from the GDPA generator, we can generate $p$ directly by adjusting the output size of pattern decoder $G_P$ to $3\times w\times h$. Directly generating $p$ can simplify the pipeline of GDPA as we do not need to translate $pattern$ to generate $p$ in two steps. Thus, it's worth investigating which design choice works better. Table~\ref{tab:p_delta} shows the results comparing these two design choices. As we can see, generating $pattern$ achieves significantly higher ASRs than generating $p$ directly. We conjecture that this is because $p$ has a larger space to optimize than $pattern$, and thus is more difficult to optimize. Hence, in our GDPA pipeline we generate $pattern$ first and then translate $pattern$ to generate $p$.

\begin{table}[ht!]
  \caption{ASRs of GDPA when generating $pattern$ vs. $p$.}\label{tab:p_delta}
  \begin{center}
  \begin{tabular}{lcc}
  \toprule
  & Generate $pattern$ & Generate $p$ \\ 
  \midrule
  Traffic Sign & \textbf{87.9\%}           & 69.7\%              \\ 
  VGGFace      & \textbf{46.3\%}           & 24.9\%              \\ 
  ImageNet     & \textbf{96.3\%}           & 63.8\%              \\ 
  \bottomrule
  \end{tabular}
  \end{center}
  \end{table}
  
\subsection{Visibility $\alpha$ vs. ASR}

\begin{figure}[ht!]
	\begin{center}
		\includegraphics[width=0.6\linewidth]{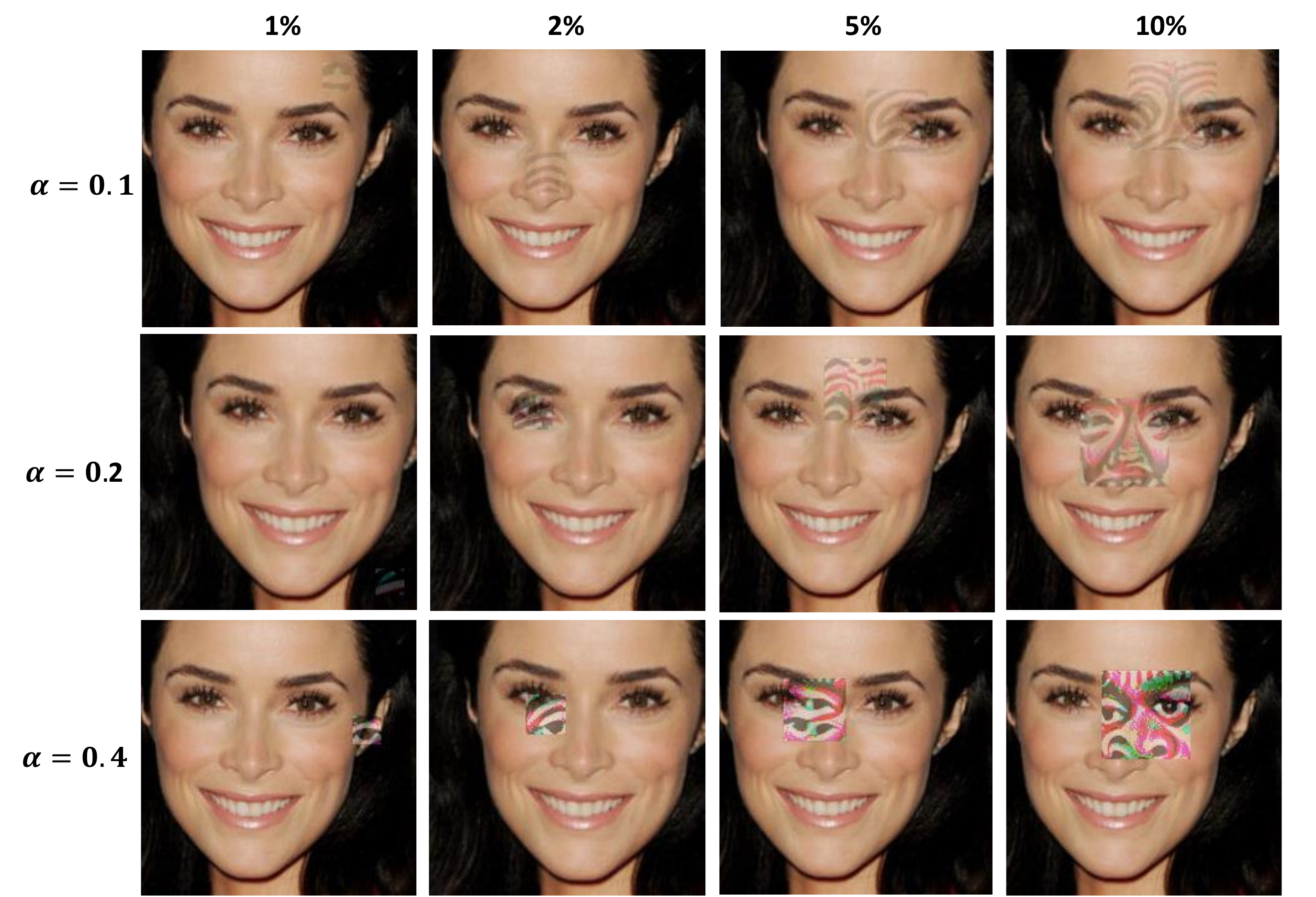}
	\end{center}\vspace{-10pt}
  \caption{Perturbed images generated by GDPA with different $\alpha$'s and patch sizes (1\%, 2\%, 5\% or 10\% pixels).}
	\label{fig:vis_invisible}
\end{figure}

In Section 4.1, we investigate the impact of visibility parameter $\alpha$ of Eq. 6 on GDPA’s ASR. Figure~\ref{fig:vis_invisible} visualizes some example perturbed images generated by GDPA with different $\alpha$'s and patch sizes. As we can see, by using different $\alpha$'s, we can control the visibility of GDPA attack. 

\subsection{Effect of $\beta$}
\begin{figure}
\begin{floatrow}
\ffigbox[\FBwidth]{
  \includegraphics[width=0.9\linewidth]{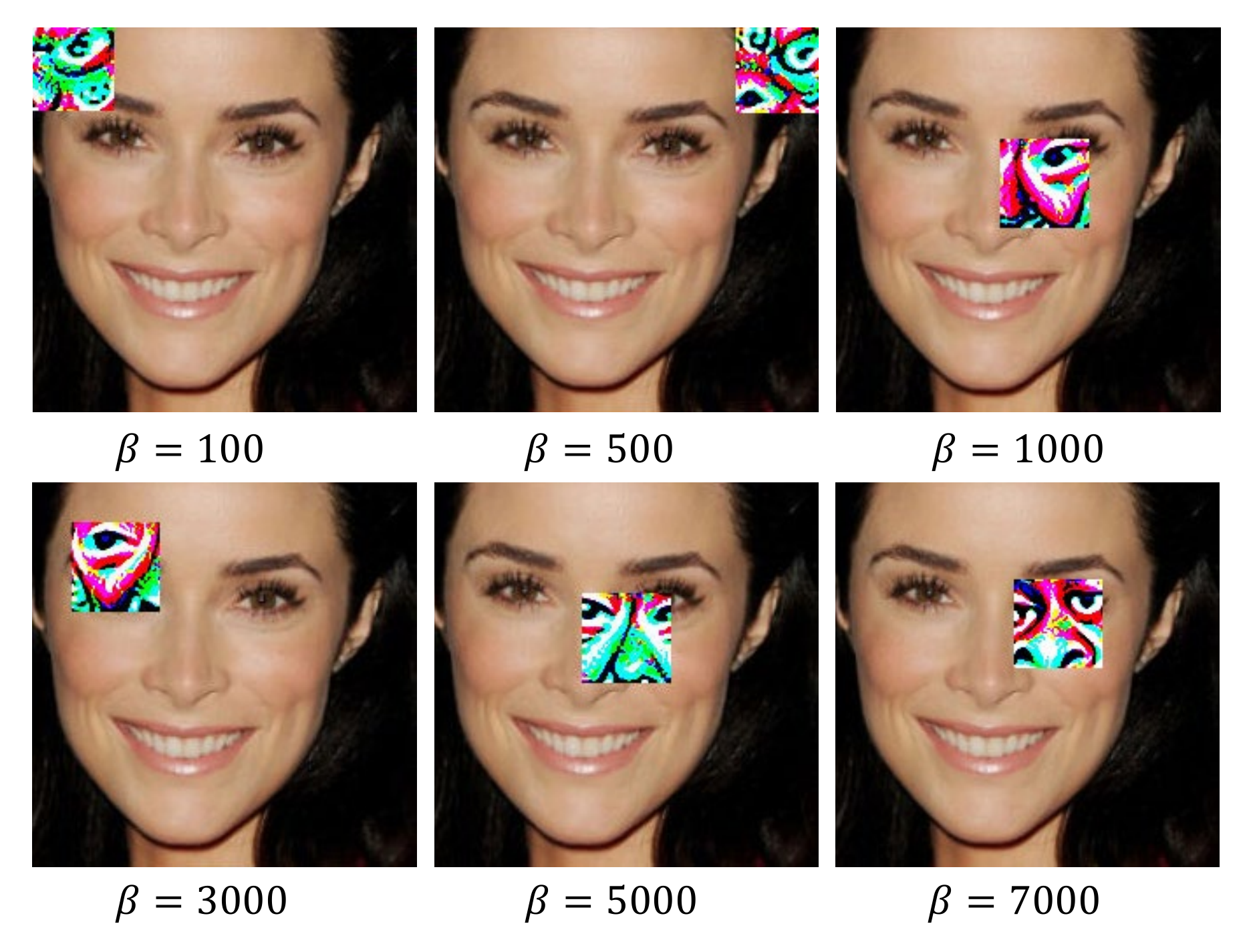}
}{
    \caption{Perturbed images by GDPA with different values of $\beta$. The patch size is 5\% of pixels.} 
	\label{fig:vis_beta}
}

\ffigbox[\FBwidth]{
  \includegraphics[width=0.9\linewidth]{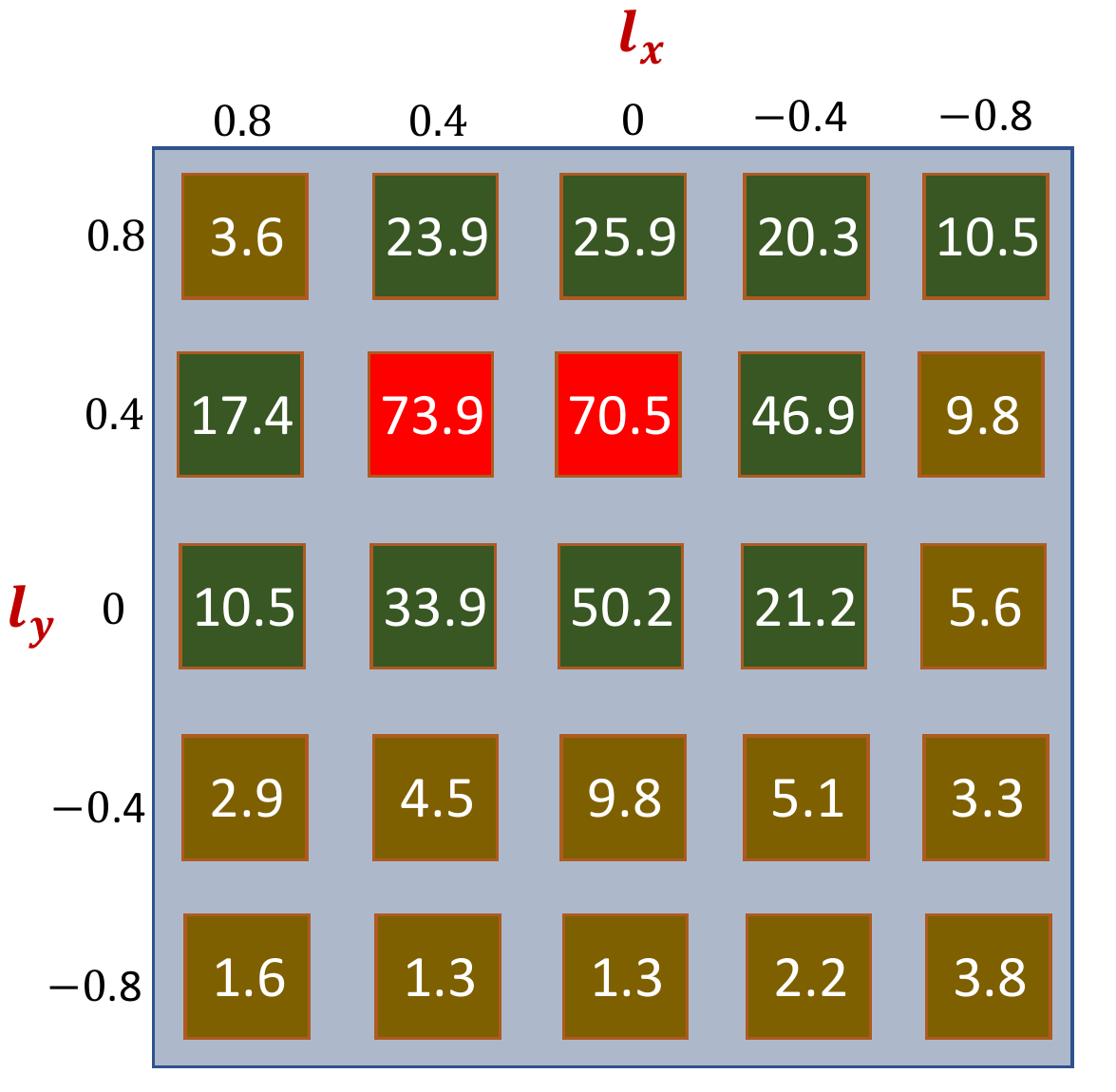}
}{
  \caption{ASRs of static patch attack on different locations. We use different colors to denote ASRs in different ranges. Red: above 70\%; Green: 10\% - 70\%; brown: below 10\%. Dynamic GDPA achieves 76.4\% ASR in this experiment.}
	\label{fig:static_patchs}
}

\end{floatrow}
\end{figure}

The $\beta$ in Eq. 1 controls the slope of $\tanh$ that constrains $l_x$ and $l_y$ in the range of $[-1, 1]$. It is critical to find an appropriate value of $\beta$ to train the GDPA generator. Intuitively, a too large or too small $\beta$ value can cause different training difficulties. If $\beta$'s value is too small, the $\tanh$ activation function saturates quickly and pushes $l_x$ and $l_y$ to the saturated value of -1 or 1, which corresponds to corners of an image. On the other hand, if $\beta$ is too large, the $\tanh$ activation function has a slow transition from -1 to 1, which may not be able to push $l_x$ and $l_y$ away from the origin $[0,0]$ of an image, and likely causes ineffective training as well. Therefore, we treat $\beta$ as a hyperparameter and tune it on the validation set. The results with different values of $\beta$ on VGGFace are shown in Table~\ref{tab:asr_beta} and Figure~\ref{fig:vis_beta}. It can be observed that we get the highest ASR with $\beta = 3000$. With small $\beta$s like $100$ or $500$, the patch location saturates at the corners of images; With large $\beta$s such as $5000$ or $7000$, the learned patch locations are close to the origin for most of the images. We find $\beta=3000$ works well across a variety of architectures and datasets, and thus set it as the default value.

\begin{table}[ht!]
\caption{ASRs of GDPA with different values of $\beta$. We use 5\% of pixels as the patch size.}\label{tab:asr_beta}
\begin{center}
\begin{tabular}{lcccccc}
  \toprule
  $\beta$ & 100 & 500 & 1000 & 3000 & 5000 & 7000 \\
  \midrule
   ASR & 15.6 & 20.8 & 88.2 & \textbf{88.4} & 88.1 & 87.9 \\
  \bottomrule
\end{tabular}
\end{center}
\end{table}

\section{Eyeglass Attack Visualization}

\begin{figure}[ht!]
\centering
\includegraphics[width=0.7\linewidth]{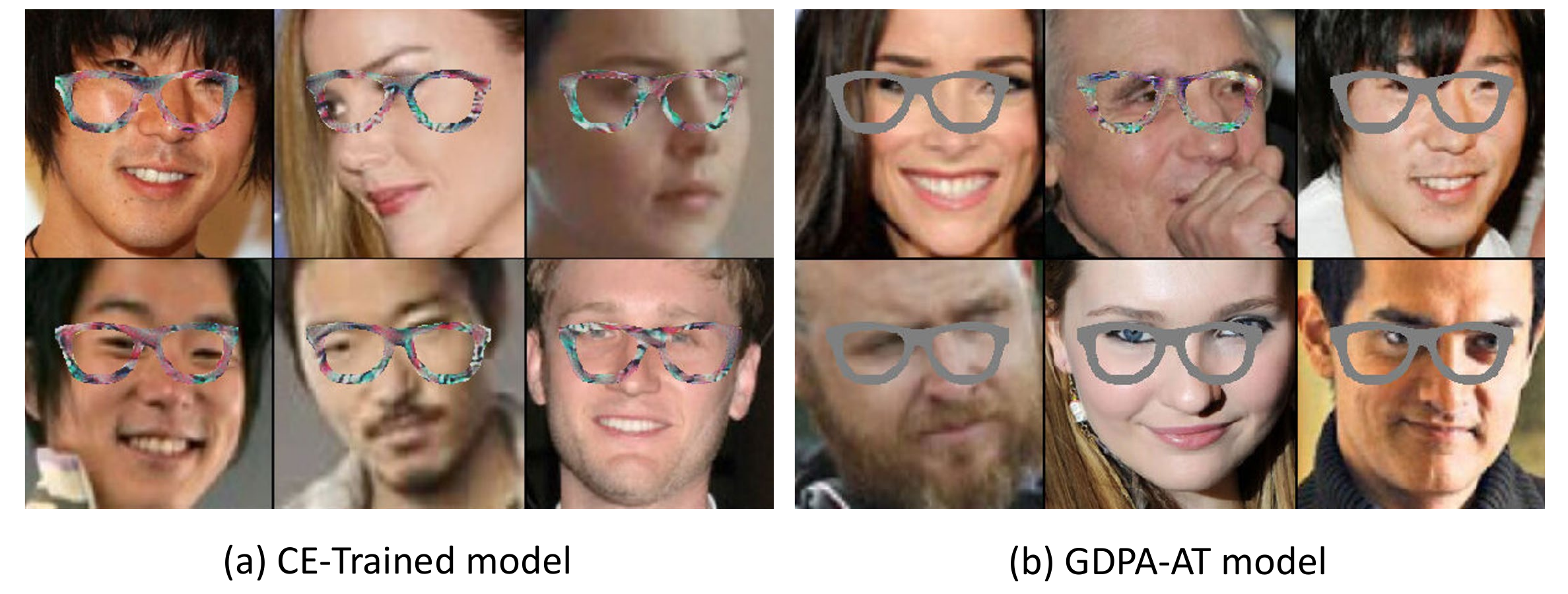}
\caption{Perturbed images generated by eyeglasses attack on (a) standard CE-trained model, and (b) GDPA-AT trained model.}
\label{fig:at_glass_vis}
\end{figure}

Figure~\ref{fig:at_glass_vis} shows example results when using eyeglasses attack to evade a standard CE-trained model~(a) and the GDPA-AT trained model~(b). As we can see, the eyeglasses attack fails to attack the GDPA-AT trained model because it is not able to generate effective adversarial patterns on the eyeglass frames in 5 out of 6 cases, while being very successful on standard CE-trained model. 

\section{Generating Static Patch Attack with GDPA}

Contrary to dynamic patch attack, static patch attack uses a fixed patch location for all the images. To conduct static patch attack with GDPA, we set $l_x$ and $l_y$ to fix values instead of generating them from $G_L$. To compare the performance between dynamic and static patch attacks, we conduct static patch attacks on VGGFace at 25 fixed locations ($l_x, l_y\in[-0.8, -0.4, 0, 0.4, 0.8]$). We use patch size $32\times 32$ ($2\%$ of pixels) in the experiment. 

Figure~\ref{fig:static_patchs} shows the ASRs of static patch attacks at the 25 locations. As we can see, patch location is an important factor in the performance of static patch attack. Notably, patch locations around the area of eyes have the best ASRs. The highest ASR we obtain from static patch attack is 73.9\%, while dynamic GDPA achieves 76.4\%, demonstrating the effectiveness of dynamic GDPA.

\section{Generating Adversarial Attack with GDPA}

Thanks to its generic formulation, we can also generate conventional adversarial attacks with GDPA by adjusting its pipeline slightly. To do this, we use a fixed mask of value $0.5$ for all image pixels, and update the generator to produce $p$ of the same size of image directly. To make sure the adversarial noise is within a small $L_\infty$-norm bound,  we multiple $p$ by $\epsilon/255$ such that the adversarial noise is bounded by $\epsilon/255$. Finally, we scale the perturbed image by 2 and clip its pixel values to $[0, 1]$ to create an adversarial example. We call this GDPA version of adversarial examples as GDPA-ADV.

\begin{figure}
\begin{floatrow}
\ffigbox[\FBwidth]{
  \includegraphics[width=0.8\linewidth]{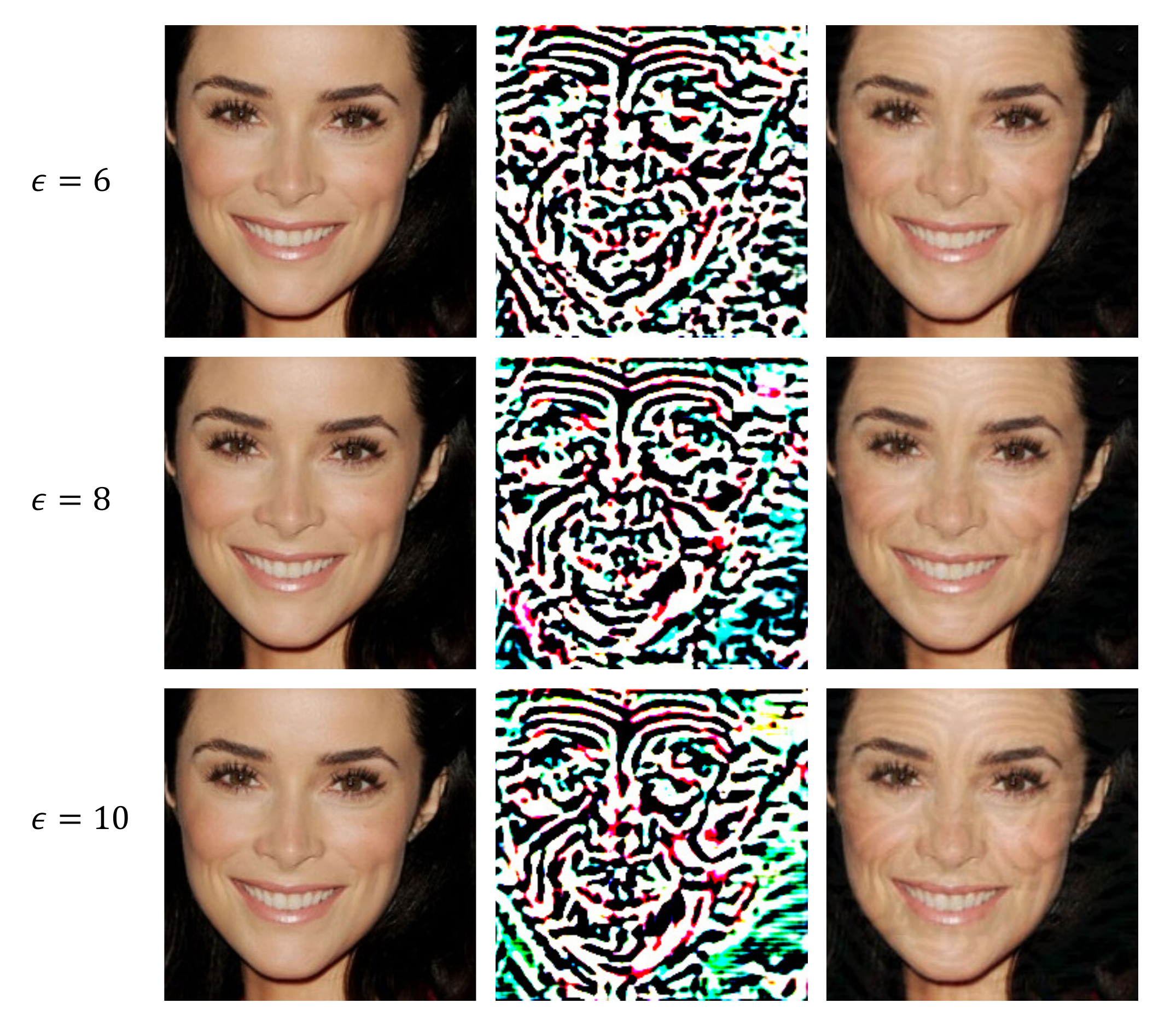}
}{
  \caption{Adversarial examples generated by GDPA-ADV with different $\epsilon$'s. Left: original images; Middle: adversarial noise scaled to $[0, 1]$ for visualization; Right: adversarial examples.}\label{fig:gdpa_adv}
}

\ffigbox[\FBwidth]{
  \includegraphics[width=0.9\linewidth]{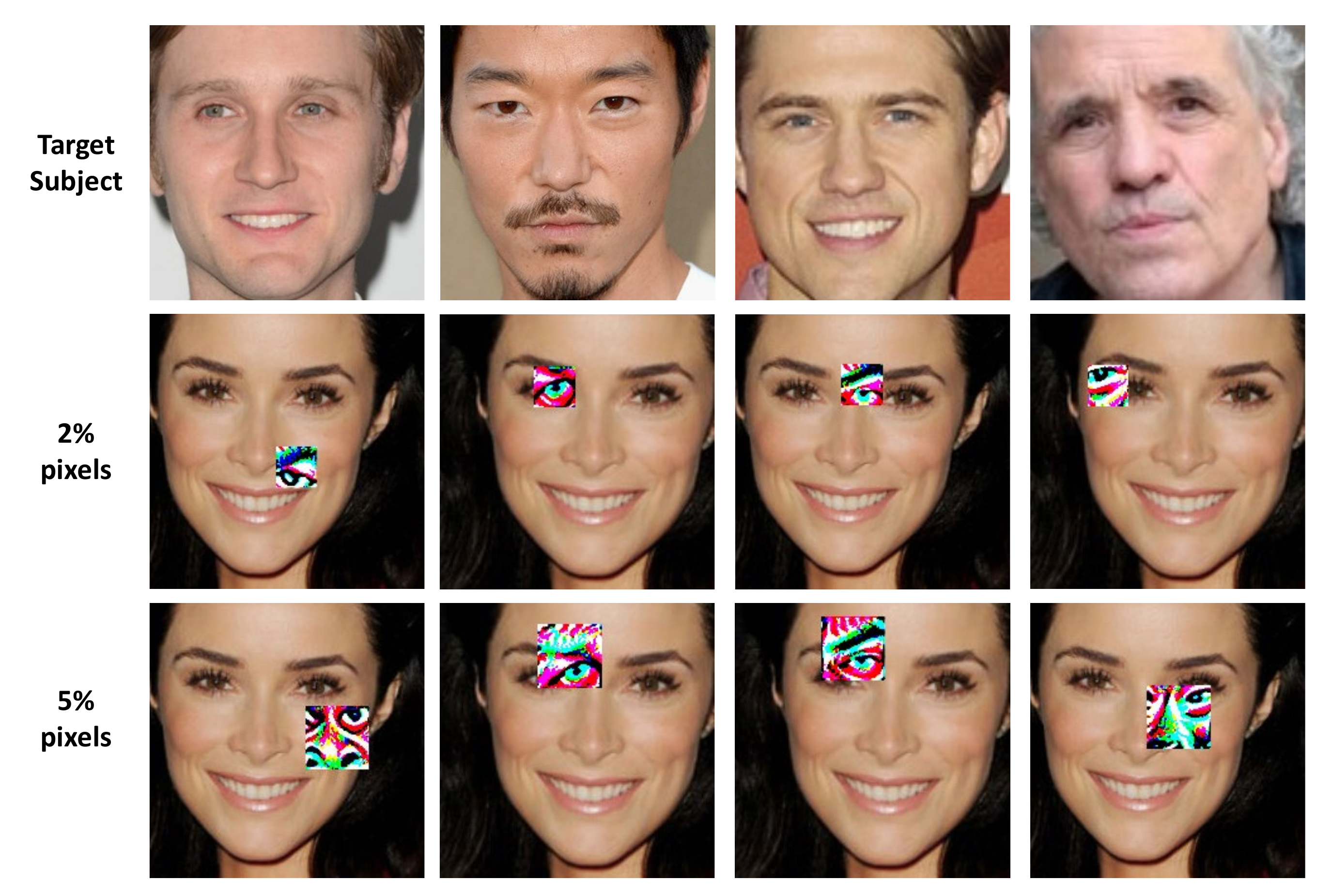}
}{
  \caption{Perturbed images generated by GDPA with targeted attack on VGGFace. Each column corresponds to one targeted attack with a different target subject.}\label{fig:vis_targ_more}
}

\end{floatrow}
\end{figure}

We then compare the attack performances of GDPA-ADV with PGD~\cite{madry2017towards} and PI-FGSM~\cite{gao2020patch} on VGGFace and ImageNet. The PGD attack is generated with learning rate 10 for 20 iterations. The results with different $\epsilon$'s are provided in Table~\ref{tab:gdpa_adv}. We can observe that GDPA-ADV achieves slightly higher ASRs than PGD in all the cases considered. Compared with the other more competitive method PI-FGSM, GDPA-ADV has slightly worse ASR except on VGGFace when $\epsilon=6$. Some adversarial examples generated by GDPA-ADV on VGGFace are visualized in Figure~\ref{fig:gdpa_adv}. These adversarial examples look similar to the conventional adversarial examples.

\begin{table}[h!]
\caption{ASRs of the adversarial attacks generated by PGD, PI-FGSM and GDPA-ADV.}\label{tab:gdpa_adv}
\begin{center}
\begin{tabular}{llccc}
  \toprule
  & & $\epsilon=6$ & $\epsilon=8$  & $\epsilon=10$ \\ 
  \midrule
  \multirow{3}{*}{VGGFace}  & PGD               & 81.5  & 90.7  & 97.8  \\
  & PI-FGSM & 79.7  & \textbf{94.9}  & \textbf{100}  \\
   & GDPA-ADV & \textbf{82.6}  & 91.9  & 98.2 \\ 
  \midrule
  \multirow{3}{*}{ImageNet} & PGD               & 88.5  & 90.3  & 91.2  \\
  & PI-FGSM & \textbf{92.9}  & \textbf{94.0}  & \textbf{94.8}  \\
    & GDPA-ADV & 89.3  & 93.2  & 94.3  \\ 
  \bottomrule
\end{tabular}
\end{center}
\end{table}

\section{GDPA-AT against Adversarial Attack}


We evaluate the robustness of models under conventional adversarial attacks, such as the PGD attack~\cite{madry2017towards}. The results are reported in Table~\ref{tab:pgd}, where different PGD attack strengths $\epsilon$ have been considered. We set step size as 20 and iterations as 300. It can be observed that GDPA-AT achieves significantly higher robustness than DOA against the PGD attack. More interestingly, the accuacies that GDPA-AT achieve are almost on par with PGD-AT even though GDPA is a patch attack algorithm. We believe this is because during the adversarial training process, GPDA generates the adversarial patches to attack the classifier iteratively; even though each patch attack is localized, the combination of all patch attacks generated during the iterative process resembles a whole image attack that PGD usually produces. For this reason, the model trained by GDPA-AT can defend conventional adversarial attacks. 


These results demonstrate that GDPA-AT is a generic defense algorithm that can defend both patch attacks and conventional adversarial attacks, while PGD-AT and DOA fail on one of them.

\begin{table}[ht!]
\begin{center}
{\small
\begin{tabular}{lcccccccccc}
\toprule
& \multicolumn{5}{c}{VGG Face} & \multicolumn{5}{c}{Traffic Sign} \\ 
\cmidrule(lr){2-6} \cmidrule(lr){7-11}
Attack Strength ($\epsilon$) & 0        & 2       & 4      & 8   & 16  & 0        & 2       & 4      & 8   & 16        \\ 
\midrule
CE training & 98.9    & 44.4   & 1.7    & 0   & 0  & 98.7    & 89.5   & 61.6     & 24.6   & 5.1    \\ 
PGD-AT~\cite{madry2017towards}  & 97.3 & 96.9 & 96.6 & 96.1    & 95.8  & 97.5 & 95.8 & 94.6 & 92.9     & 91.0 \\ 
\midrule
DOA-Grad~\cite{wu2019defending}         & 97.5    & 33.4   & 0.4   & 0   & 0           & 95.6 & 91.2 & 79.5 & 46.9 & 6.7    \\ 
DOA-Exh~\cite{wu2019defending}         & 98.5    & 35.7   & 0.4   & 0   & 0      & 92.9 & 89.5 & 77.1 & 42.8 &    5.8   \\ 
\textbf{GDPA-AT}         & \textbf{98.9}    & \textbf{95.1}    & \textbf{94.9}   & \textbf{94.6}    & \textbf{94.5}  & \textbf{98.5} & \textbf{94.7}   & \textbf{93.5}    & \textbf{92.2}  &   \textbf{90.3} \\ 
\bottomrule
\end{tabular}
}
\end{center}
\caption{The accuracies of different robust models on (a) VGGFace, and (b) Traffic Sign when under the PGD attack.}
\label{tab:pgd}
\end{table}

\section{Cross Attacks and Defenses}
In this section, we compare the defense performances of PGD-AT, DOA and GDPA-AT when they are attacked by their corresponding attack algorithms. In this experiment, the PGD attack uses $\epsilon=8$, and ROA and GDPA use 10\% pixels as patch size. The results on VGGFace are shown in Table~\ref{tab:at_res}. As we can see, PGD-AT achieves the highest robustness under the PGD attack, but is not very robust under the ROA and GDPA attacks. On the other hand, DOA achieves decent robustness under the ROA and GDPA attacks, but fails completely under the PGD attack. Notably, GDPA-AT is the only defense algorithm that achieves almost the highest robustness under all three attacks. It's expected that GPDA-AT would be robust under the ROA and GDPA attacks since both are patch attacks. An explanation of the robustness of GDPA-AT under the PGD attack is provided in Section 4.2. 

\begin{table}[ht!]
\caption{Accuracies of adversarially trained models under PGD, ROA and GDPA attacks.}
\label{tab:at_res}
\begin{center}
\begin{tabular}{l|ccc}
  \toprule
  \diagbox{AT}{Attack} & PGD & ROA & GDPA \\ 
  \midrule
  PGD-AT    & \textbf{96.1}  & 32.8  & 30.5  \\
  DOA  & 0     & 88.1  & 86.9  \\
  GDPA-AT   & 94.6  & \textbf{90.4}  & \textbf{88.2}  \\ 
  \bottomrule
\end{tabular}
\end{center}
\end{table}

\section{Additional Results on Targeted Attack}

Figure~\ref{fig:vis_targ_more} provides additional perturbed images generated by targeted GDPA attack on VGGFace. The top row shows the target subjects, while the bottom two rows show the perturbed images with different patch sizes. As we can see, the patches generated by GDPA attempt to replace the corresponding face features with the ones from the target subjects.

\end{document}